%% file: main.tex
\documentclass[letterpaper, 10pt, conference]{ieeeconf}

\IEEEoverridecommandlockouts %
\usepackage{xspace}
\usepackage{caption}
\usepackage{subcaption}
\usepackage[ruled,linesnumbered]{algorithm2e}
\usepackage{url}
\usepackage{booktabs}
\usepackage{graphicx}
\usepackage{float}
\usepackage{placeins}
\usepackage{multirow}
\usepackage{tabularx}
\usepackage{makecell}
\let\labelindent\relax %
\usepackage{enumitem}
\usepackage{diagbox}
\usepackage{amsmath}
\usepackage{amssymb}
\usepackage{wrapfig}
\usepackage{pifont} %
\usepackage{array}

\usepackage{xcolor}
\makeatletter\let\NAT@parse\undefined\makeatother %
\usepackage[square,numbers]{natbib}
\usepackage{hyperref}

\newcommand{\sysname}{\textit{\textsc{ViFiNav}\xspace}}
\newcommand{\simname}{\textit{\textsc{WiFiSpaces}\xspace}}

\definecolor{responseblue}{RGB}{66, 109, 181}

\providecommand{\keywords}[1]{}

\title{\LARGE \bf Visual--Wi-Fi Navigation to Signal-Emitting Devices}

\author{Lihao Wang, Tianjian Xu, Tianmin Shu, and Renjie Zhao%
\thanks{Johns Hopkins University, United States.}%
\thanks{T. Xu contributed to this work while a visiting student at Johns Hopkins University.}%
}

\begin{document}

\maketitle
\bstctlcite{IEEEexample:BSTcontrol}  
\thispagestyle{empty}
\pagestyle{empty}

\input{Latex_Files/abstract}

\input{Latex_Files/intro}

\input{Latex_Files/related}

\input{Latex_Files/methods}

\input{Latex_Files/implementation}

\input{Latex_Files/experiments}
\input{Latex_Files/conclusion}

\bibliographystyle{IEEEtranN}
\bibliography{ref}

\end{document}

%% file: Latex_Files/abstract.tex
\begin{abstract}
Indoor navigation to task-relevant targets remains difficult when the target is small, occluded, or visually indistinct.
Meanwhile, smart appliances, mobile devices, and IoT tags emit Wi-Fi signals during everyday communication, providing a pervasive but underused spatial cue.
We study visual--Wi-Fi navigation, where a robot reaches a signal-emitting device in an unmapped environment using onboard visual observations and Wi-Fi packets overheard from the device, without connecting to the device or modifying infrastructure.
Realizing this requires pose-dependent Wi-Fi observations during closed-loop policy learning and reliable spatial cues from bearing measurements corrupted by indoor multipath propagation on a moving robot.
We present \simname{}, which extends the Habitat embodied simulator with online ray-traced Wi-Fi sensing, rendering Wi-Fi observations for a robot-mounted antenna array at continuous poses during policy rollouts.
On top of it, we build \sysname{}, a localization-aware navigation framework: an online source-localization algorithm exploits robot motion to disambiguate and robustly aggregate bearings from multiple viewpoints into a confidence-aware source estimate, and a learned policy fuses this evolving estimate with depth observations and learns to act on it as it is refined.
Experiments show that exploiting robot motion is key to accurate source estimates, and that fine-tuning on ray-traced Wi-Fi observations in \simname{} improves the success rate of \sysname{}. The policy transfers zero-shot to unseen scene datasets, and we deploy it on a real robot with a Wi-Fi sensor built from a commodity router, without any real-world training.

\end{abstract}

%% file: Latex_Files/intro.tex
\section{Introduction}\label{sec:intro}
\begin{figure}[t]
    \centering
    \begin{subfigure}[b]{0.680\linewidth}
        \centering
        \includegraphics[width=\linewidth]{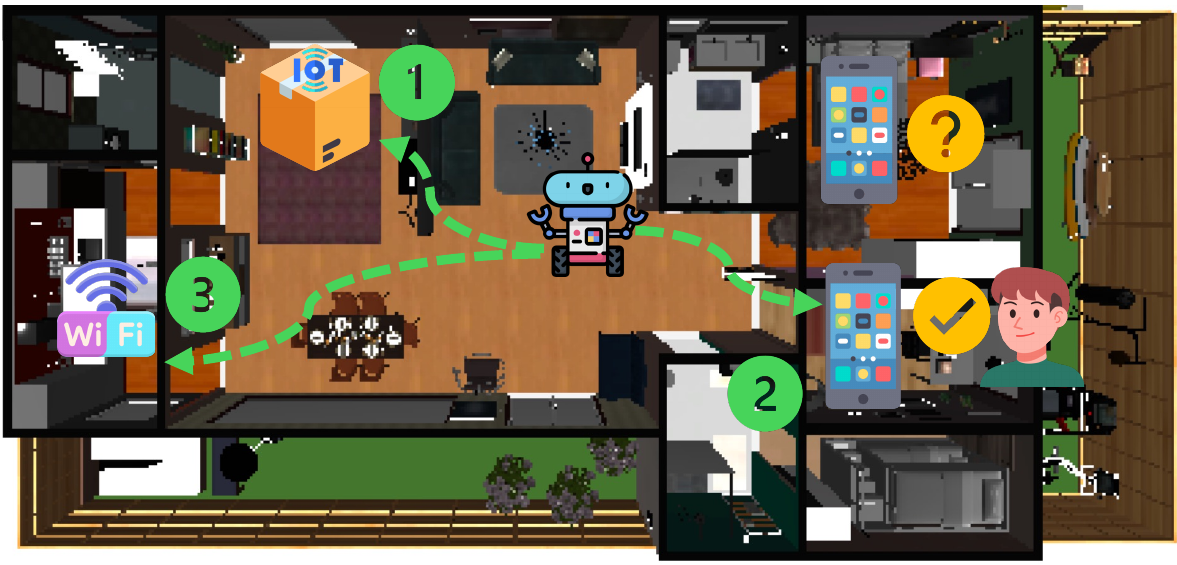}
        \caption{Task.}
        \label{fig:task_scene}
    \end{subfigure}
    \hfill
    \begin{subfigure}[b]{0.295\linewidth}
        \centering
        \includegraphics[width=\linewidth]{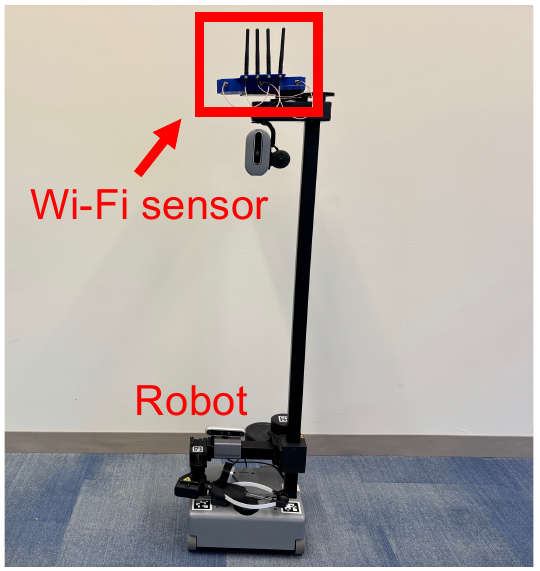}
        \caption{Robot.}
        \label{fig:hardware_robot}
    \end{subfigure}
    \caption{Visual--Wi-Fi navigation in everyday indoor environments. (a) Using visual and Wi-Fi observations, an agent navigates toward signal-emitting devices in an unmapped indoor environment. Examples: 1. finding objects equipped with signal-emitting IoT tags (e.g., a package with an AirTag-style Wi-Fi tag), 2. finding the user with a carried mobile device, and 3. finding non-connected objects via associated signal-emitting devices (e.g., food in a smart fridge). (b) Our robot: a commodity Wi-Fi router mounted on a Hello Robot Stretch~3 serves as the Wi-Fi sensor.\label{fig:task}}
\end{figure}

Navigation in unmapped environments often involves locating and reaching task-relevant targets, including everyday connected devices such as smart appliances and mobile devices.
While visual navigation has made substantial progress~\citep{wijmans2022ver,kim2023topological,khanna2024habitat}, it remains difficult when the target is small, occluded, visually indistinct, or has no direct visual manifestation~\citep{young2025enhancing}. A vision-only agent then lacks an explicit spatial cue and must rely on weak semantic priors or broad exploration, producing inefficient trajectories.
Meanwhile, the growing prevalence of connected devices creates a new sensing opportunity: many indoor targets already emit wireless signals as part of their normal communication, which can be overheard by a robot-mounted receiver to provide complementary spatial cues.
As shown in Fig.~\ref{fig:task}, we study \textit{visual--Wi-Fi navigation}: in an unmapped environment, an embodied agent uses onboard visual observations and Wi-Fi packets overheard from the target's existing traffic to navigate to the physical location of a signal-emitting target device. The target is defined by the Wi-Fi source location and need not be visible.

Unlike audio cues, which require the target to actively and continuously emit informative sound and are susceptible to environmental noise~\citep{chen2020soundspaces,chen2022soundspaces,shi2025towards}, wireless signals are emitted by ordinary devices as a by-product of communication and propagate through and around obstacles~\citep{hunter2023wiros,mundra2024wisensehub}, providing spatial cues beyond visual line of sight.
Among wireless technologies, we focus on Wi-Fi because it is the most pervasive: ultra-wideband (UWB) and Bluetooth Low Energy (BLE) localization typically require dedicated anchors or tags to be deployed and calibrated~\citep{grosswindhager2017uwb,zhao2021uloc,zhang2025bridge}, whereas most household and personal devices, including IoT tags, already communicate over Wi-Fi (Fig.~\ref{fig:task}), and their packets can be overheard by a commodity Wi-Fi router without connecting to the device or its network~\citep{gringoli2019free,hunter2023wiros}.
Wi-Fi is a deployment choice, not a requirement: the same setting arises for any wireless modality that carries spatial cues beyond line of sight.
However, autonomous navigation with Wi-Fi signals raises two key challenges.

The first challenge is to obtain pose-dependent Wi-Fi observations at scale for learning-based embodied navigation: during rollouts they must track the agent's continuous pose, including heading, while reflecting the source location and the 3D geometry that shapes propagation, reflection, and blockage.
Precomputed radio maps do not fit: they tabulate the signal on a discrete grid and would have to be rebuilt every episode as the source moves, over poses that are policy-dependent and unknown in advance.
To address this, we propose \simname{}, to our knowledge the first embodied navigation simulator to provide online Wi-Fi observations at the agent's current pose, by extending Habitat with a Wi-Fi sensor built on physics-based ray tracing~\citep{sionna}; synchronizing agent state with the wireless scene at every step and reusing scene geometry across episodes makes online ray tracing practical for closed-loop policy learning.

The second challenge is to turn noisy Wi-Fi measurements into reliable spatial cues on a moving robot. Each overheard packet yields channel state information (CSI) at the router, from which a bearing to the source can be estimated~\citep{kotaru2015spotfi,hunter2023wiros}, but indoor multipath makes single-vantage bearings heavy-tailed: in \simname{}, about $10\%$ of raw bearings are off by more than $30^\circ$ (Fig.~\ref{fig:loc_analysis_a}), consistent with real deployments~\citep{kotaru2015spotfi,xie2019md,eleftherakis2024spring+}, and acting on such an estimate commits the robot to the wrong part of the scene. 
To address this, we present \sysname{}, a localization-aware visual--Wi-Fi navigation framework: an online Wi-Fi source localization algorithm resolves the directional ambiguity of the small antenna array using the robot's natural motion and robustly aggregates multi-view bearings into a confidence-aware source estimate, and, rather than heuristically following each intermediate estimate, a learned multimodal policy fuses this evolving cue with visual observations and learns to act on it in closed loop.

In summary, our contributions are as follows:
\begin{itemize}[leftmargin=*,itemsep=0pt,topsep=1pt,parsep=0pt]
\item A complete system for visual--Wi-Fi navigation, which we formulate as reaching a signal-emitting device in an unmapped environment from visual observations, odometry, and Wi-Fi packets overheard from its existing traffic, without modifying the target device or infrastructure.
\item \simname{}, an embodied navigation simulator that provides online ray-traced Wi-Fi observations synchronized with continuous agent poses during policy rollouts.
\item An online Wi-Fi source localization algorithm coupled with a reinforcement-learned policy that learns to act on its evolving estimates.
\item Simulation evaluation and real-world deployment: fine-tuning in \simname{} improves the success rate of \sysname{} over the same policy without Wi-Fi fine-tuning, and the policy generalizes zero-shot across indoor scene datasets; the same policy is further deployed on a real robot with a commodity-router Wi-Fi sensor~\citep{gringoli2019free,hunter2023wiros}, without any real-world training.
\end{itemize}

%% file: Latex_Files/related.tex
\section{Related Work}

\textbf{Embodied navigation with vision and sound.}
Visual navigation has been widely studied, including PointGoal \citep{savva2019habitat,wijmansdd,ye2021auxiliary,wijmans2022ver}, ObjectGoal \citep{chaplot2020object,khanna2024habitat,ko2025context}, ImageGoal \citep{kim2023topological}, and vision-language navigation \citep{long2025instructnav}, but relies on target visibility or learned semantic priors and degrades to weakly guided exploration when the target is small, occluded, or has no visual manifestation.
Audio-visual navigation~\citep{chen2020soundspaces,chen2022soundspaces,shi2025towards} adds sound cues, but requires the target to actively and continuously emit informative sound.
Wi-Fi signals, in contrast, are emitted by ordinary devices as a by-product of their normal traffic and provide spatial cues beyond visibility.

\textbf{Wireless localization and navigation.}
Indoor localization with Wi-Fi \cite{biswas2010wifi,xiong2013arraytrack,kotaru2015spotfi,xie2019md,ayyalasomayajula2020deep,jiokeng2020ftm,eleftherakis2024spring+}, UWB \cite{grosswindhager2017uwb,zhao2021uloc}, and BLE \cite{zhang2025bridge} is well studied; Wi-Fi is attractive for its commodity availability, but existing systems rely on site-specific data \cite{biswas2010wifi,ayyalasomayajula2020deep} or calibrated anchors \cite{xiong2013arraytrack,kotaru2015spotfi,xie2019md,jiokeng2020ftm,eleftherakis2024spring+}, limiting map-free use.
Robot motion has been exploited to synthesize larger apertures for angle-of-arrival (AoA) estimation \cite{jadhav2022wireless}, but with active coordination with the Wi-Fi infrastructure \cite{hunter2023wiros,mundra2024wisensehub} and for localization or tracking rather than closed-loop navigation.
Classical radio source seeking~\cite{twigg2012rss,charrow2014cooperative} couples an RF estimate with a model-based controller, but does not fuse it with visual observations in a learned policy or formulate the problem as an embodied navigation task with standardized episodes and metrics.
Prior navigation works using indoor mmWave cellular links~\cite{yin2022millimeter,yin2024zero,li2025digital} need dedicated base stations and suffer severe wall attenuation, require hard-to-obtain link-state labels, and train on precomputed radio maps rather than pose-dependent observations during closed-loop execution.

\begin{table}[t]
\centering
\footnotesize
\setlength{\tabcolsep}{3pt}
\caption{Simulation platforms for visual--Wi-Fi navigation.}
\label{tab:simulator_comparison}
\begin{tabular}{@{}lcccc@{}}
\toprule
Platform
& \makecell{Nav.\\tasks}
& \makecell{Visual\\obs.}
& \makecell{Wireless\\obs.}
& \makecell{Learning\\interface} \\
\midrule
Embodied simulators~\cite{savva2019habitat,shen2021igibson,kolve2017ai2,IsaacSim}
& \checkmark & \checkmark & -- & \checkmark \\
Wireless simulators~\cite{sionna,WirelessInSite,chen2026rfdt}
& -- & -- & \checkmark & -- \\
\simname{} (ours)
& \checkmark & \checkmark & \checkmark & \checkmark \\
\bottomrule
\end{tabular}
\vspace{-8pt}
\end{table}
\textbf{Wireless-aware simulation platforms.}
As summarized in Table~\ref{tab:simulator_comparison}, embodied simulators~\cite{savva2019habitat,shen2021igibson,kolve2017ai2,IsaacSim} provide navigation tasks, visual observations, and policy-learning interfaces but do not model radio propagation, whereas wireless simulators~\cite{sionna,WirelessInSite,chen2026rfdt} model propagation with high fidelity but are not embodied environments.
Recently, SimART~\cite{yan2026simart} couples a robotics simulator with Sionna ray tracing for outdoor 6G sensing but does not target indoor navigation, and concurrent work~\cite{lu2026rfcues} trains an RF-guided navigation policy with Habitat and Sionna but evaluates it only in simulation.
\simname{} instead embeds Wi-Fi sensing in the step loop of an embodied simulator: CSI for a robot-mounted antenna array is rendered at every agent pose in indoor scenes and exposed as an observation synchronized with visual observations and actions, so policies can be trained and evaluated closed-loop.

%% file: Latex_Files/methods.tex
\section{Task Definition}\label{sec:task}
\begin{figure}[t]
    \centering
    \includegraphics[width=\linewidth]{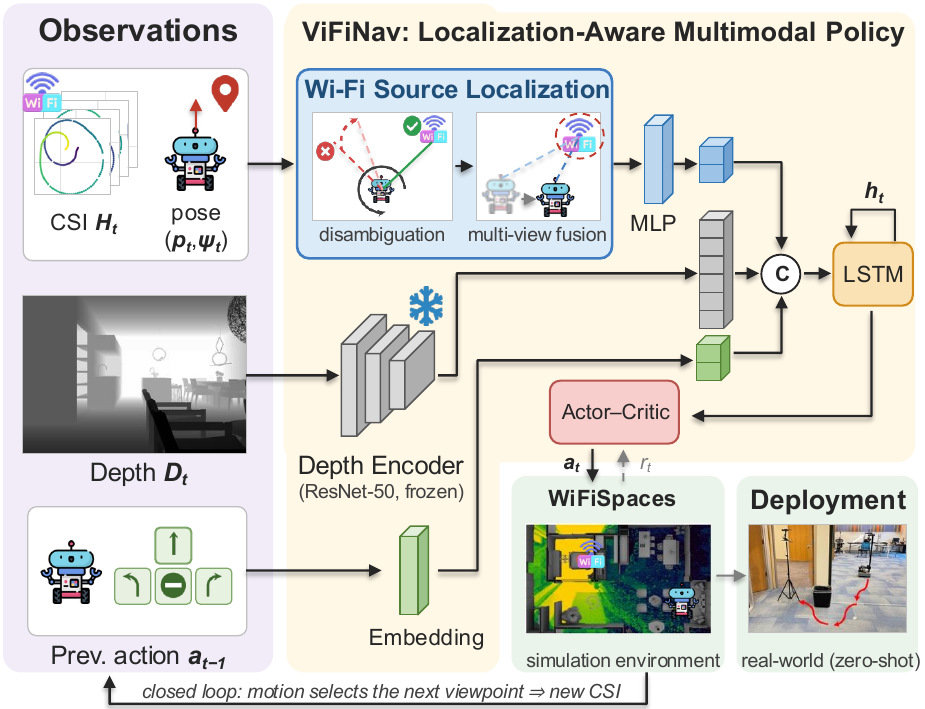}
    \caption{System overview. From CSI $\mathbf{H}_t$ and the robot pose, the online Wi-Fi source localization module of \sysname{} (disambiguation and multi-view fusion, Fig.~\ref{fig:spatial}) produces a confidence-tagged source estimate, which the policy fuses with depth and previous-action features to output the action $a_t$. They form a closed loop: the policy's motion selects the localizer's viewpoints, and the policy is trained on the evolving estimate rather than on a fixed goal. The policy is trained with reinforcement learning (RL) in \simname{}, using a reward available only in simulation, and deployed zero-shot on a real robot.\label{fig:policy}}
\end{figure}

We consider visual--Wi-Fi navigation: an agent must reach a stationary Wi-Fi-emitting device at an unknown location in an unmapped indoor scene, using only onboard observations. Unlike PointGoal and ObjectGoal navigation, neither a goal vector nor a visual description of the target is available: the target is identified only by its MAC address, and its location must be inferred from Wi-Fi measurements overheard from it, while visual observations provide the scene context needed to reach it.
At each step $t$, the agent receives $o_t=(o_t^{v}, o_t^{w})$, where $o_t^{v}$ is an egocentric visual observation (RGB-D) and $o_t^{w}$ is Wi-Fi CSI, a complex-valued channel response across antennas and orthogonal frequency-division multiplexing (OFDM) subcarriers (Sec.~\ref{sec:sim}), and its onboard odometry pose $(\mathbf{p}_t,\psi_t)$ in an episode-local frame, and acts with the standard Habitat action space (\textit{move forward}, \textit{turn left}, \textit{turn right}, \textit{stop}). An episode is specified by a scene, an initial agent pose, and a target location; the target location is used by the simulator to generate Wi-Fi observations and by the evaluator to measure success, but is never exposed to the agent. An episode succeeds if the agent executes \textit{stop} within a success distance of the source within the maximum episode length of 500 steps. Fig.~\ref{fig:policy} overviews our system.

\setcounter{topnumber}{1}%

\section{\simname{}: Extending Habitat with Continuous Wi-Fi Sensing}\label{sec:sim}

\begin{figure}[t]
    \centering
    \includegraphics[width=\linewidth]{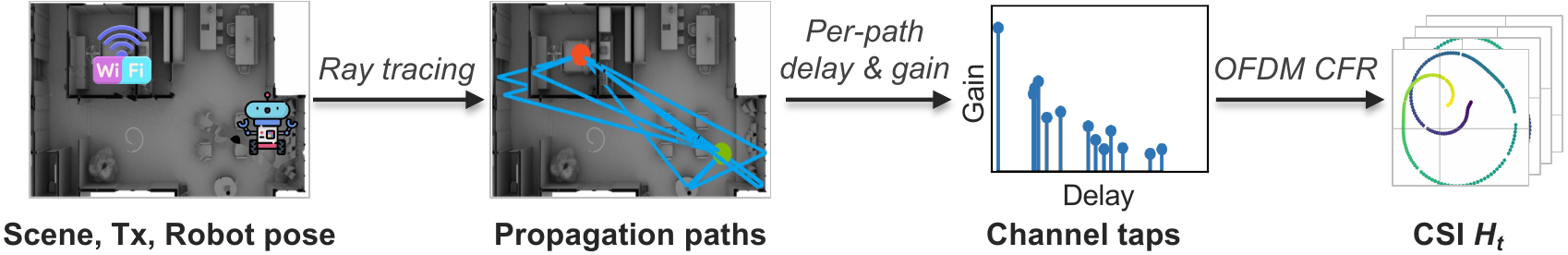}
\caption{
\simname{} renders the Wi-Fi observation on the fly. Given the scene mesh, the transmitting device and the robot pose, propagation paths, including reflections off scene surfaces, are ray-traced per antenna element; the per-path delays and gains form the channel taps, from which the Wi-Fi CSI $\mathbf{H}_t$ at the current pose is derived.
}
\label{fig:wifi_habitat}
\end{figure}

\textbf{Online ray-traced Wi-Fi sensing over continuous pose.}
We implement a Wi-Fi sensor for Habitat that renders CSI on the fly at continuous agent poses inside the Habitat step loop\footnote{We plan to release the sensor and training code upon publication.} (Fig.~\ref{fig:wifi_habitat}, Table~\ref{tab:simulator_comparison}); its propagation solver builds on the Sionna ray tracer~\cite{sionna}, which prior work has validated against indoor Wi-Fi measurements~\cite{lyu2026radiorange}.
Habitat maintains embodied navigation state, including action transitions, collisions, agent pose, and visual observations, while the Wi-Fi sensor maintains the corresponding wireless scene and ray-traces the channel.
At each policy step, the sensor takes the updated Habitat agent pose as the receiver-array pose and the episode's source position as the transmitter position, and solves for the pose-dependent propagation channel between them, including line-of-sight, specular and diffuse reflection, refraction, and diffraction; each antenna element is traced individually rather than under a far-field array approximation, so inter-antenna phase differences are geometrically consistent.
The channel response is converted into CSI across antennas and OFDM subcarriers, matching the low-level measurements exposed by commodity receivers.
The sensor is registered as a native Habitat observation, thus Wi-Fi measurements are synchronized with visual observations and actions and directly consumable by standard policy-learning pipelines.
The sensor exposes the carrier frequency, bandwidth, transmitter and receiver array layout and element spacing, and ray-tracing depth and sampling budget as configuration, which we set to match our real-world platform (Sec.~\ref{sec:impl}); the radio material of surfaces is also configurable.

\textbf{Scalable online ray tracing.}
Ray tracing needs its own scene representation: structural surfaces carrying radio materials, not the Habitat render mesh. Rebuilding it at every episode reset would repeat mesh parsing and context initialization in every parallel worker, and keeping the full render geometry resident would inflate each worker's GPU memory, limiting how many rollout workers a GPU can sustain.
\simname{} therefore separates static and dynamic wireless scene state.
Each worker holds one persistent ray-tracing context and lazily loads the current scene's static wireless mesh, built from the room-scale structural geometry of the Habitat scene; the mesh is reused across episodes and reloaded only when the scene changes.
Each observation query then supplies only the transmitter and receiver poses to the path solver on the loaded scene, never modifying the mesh; this reuse speeds up rendering without changing what is rendered.
Compared with per-query scene construction on the full scene mesh, this roughly halves both the render time of one Wi-Fi observation and the GPU memory per worker on an NVIDIA L40S GPU.

\setcounter{topnumber}{2}
\section{\sysname{}: Localization-Aware Visual--Wi-Fi Navigation}\label{sec:framework}
\subsection{Online Wi-Fi Source Localization}
\label{sec:loc}
\label{sec:wifi_loc}
\begin{figure}[t]
    \centering
    \begin{subfigure}[b]{0.48\linewidth}
        \centering
        \includegraphics[width=\linewidth]{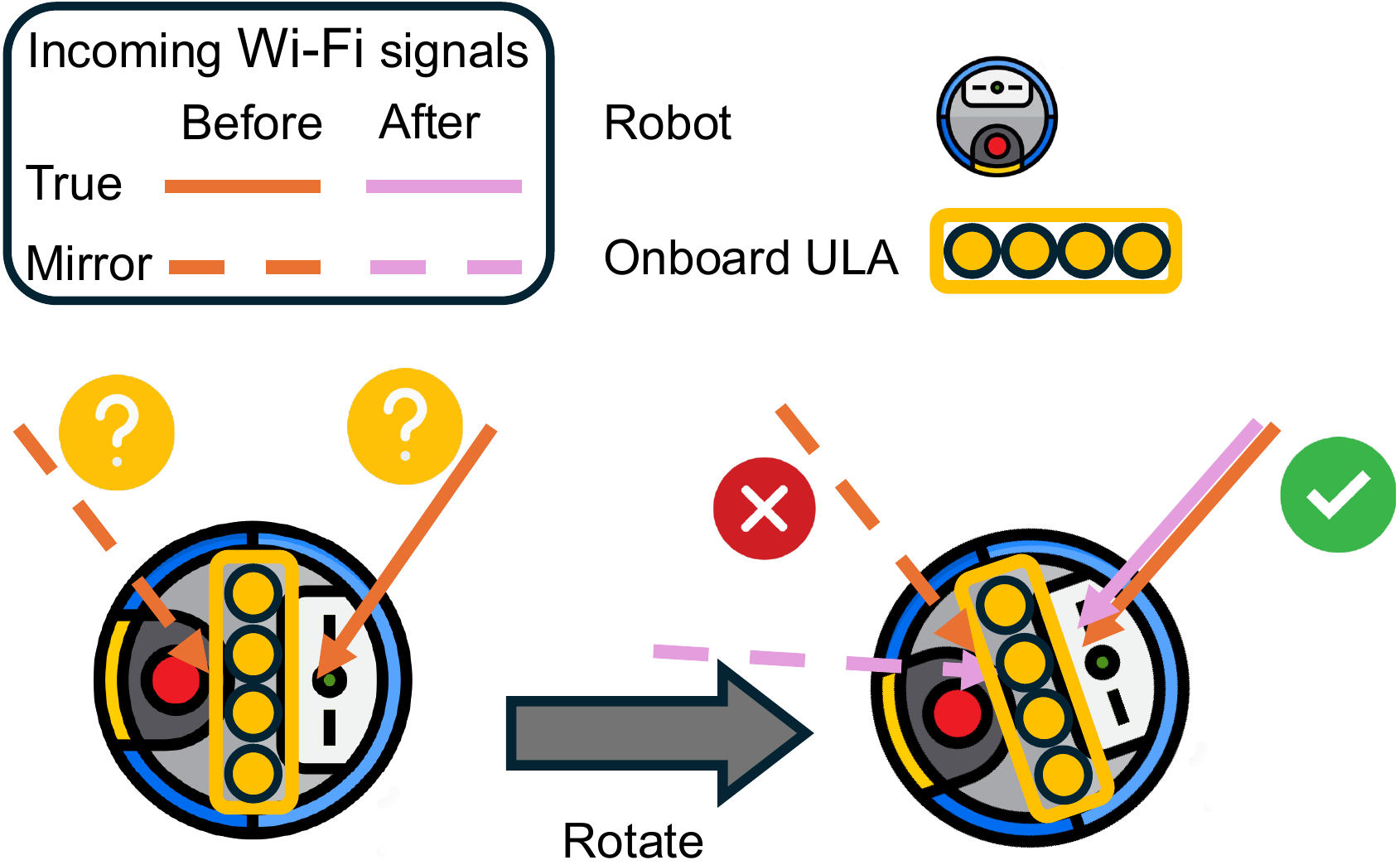}
        \caption{Disambiguation}
        \label{fig:bearsolver}
    \end{subfigure}
    \hfill
    \begin{subfigure}[b]{0.48\linewidth}
        \centering
        \includegraphics[width=\linewidth]{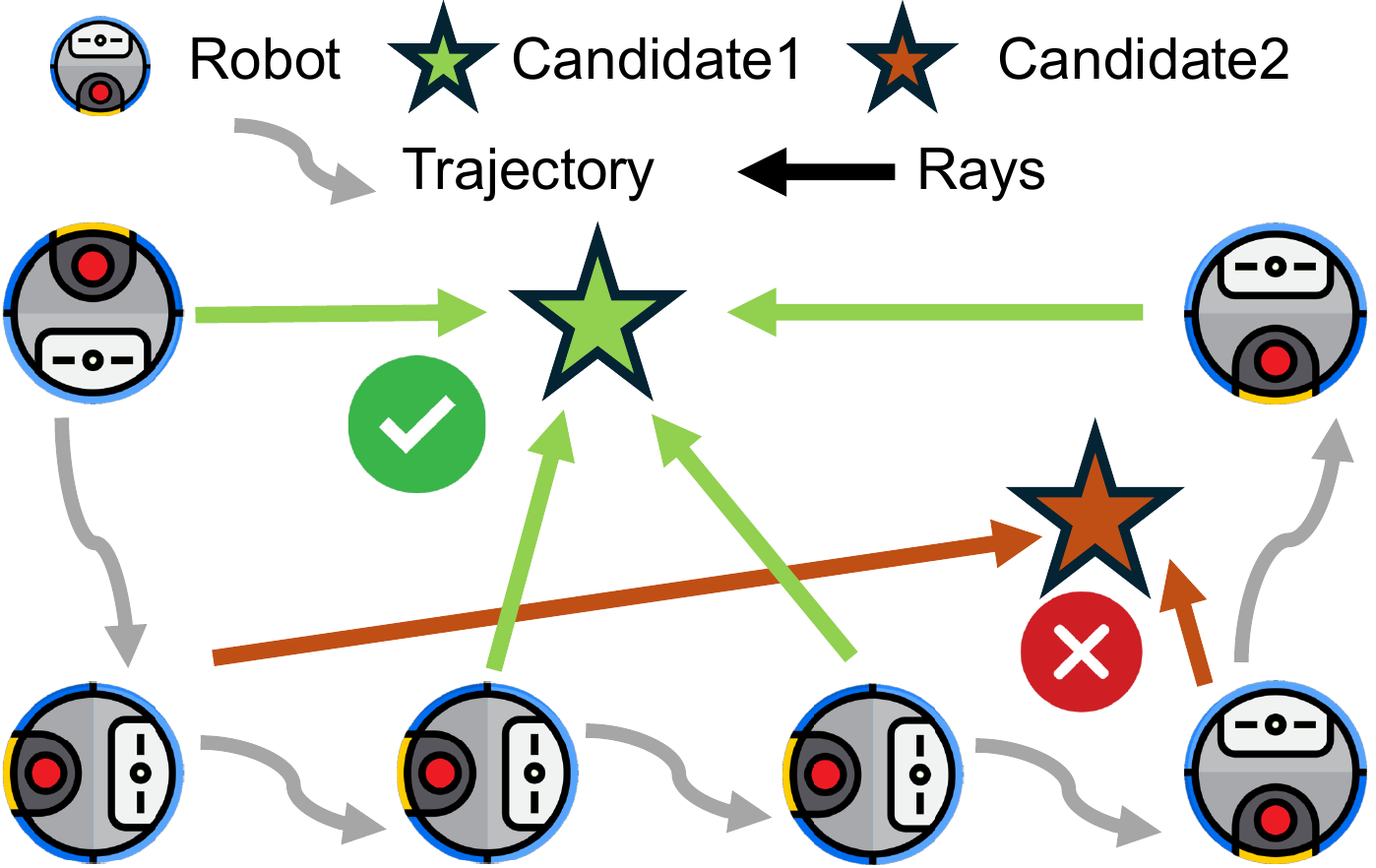}
        \caption{Multi-view fusion}
        \label{fig:ransac}
    \end{subfigure}
\caption{Wi-Fi source localization. (a) A ULA measurement yields a true and a mirror bearing; under in-place rotation only the true one stays fixed in the world frame, so it is kept. (b) Verified bearings from different viewpoints are fused by RANSAC, keeping the hypothesis with most inliers.}
    \label{fig:spatial}
\end{figure}

The localization module (Fig.~\ref{fig:spatial}, Alg.~\ref{alg:loc}) converts each CSI observation $\mathbf{H}_t\in\mathbb{C}^{N_r\times N_s}$ ($N_r{=}4$ antennas, $N_s{=}234$ subcarriers) and agent pose $(\mathbf{p}_t,\psi_t)$ into an agent-centric cue $\mathbf{z}_t^w=[\hat{\rho}_t,\cos\hat{\phi}_t,\sin\hat{\phi}_t,c_t]$, where $\hat{\rho}_t$ and $\hat{\phi}_t$ are the range and bearing of the current source estimate $\hat{\mathbf{s}}_t$ in the agent frame and $c_t\in[0,1]$ is its heuristic confidence score.

\textbf{Bearing estimation under multipath.}
Following CSI-based localization systems~\cite{kotaru2015spotfi,hunter2023wiros}, we compute a joint AoA--time-of-flight (ToF) spectrum from $\mathbf{H}_t$, in which each propagation path appears as a peak, and take the AoA $\theta_t$ of the strongest peak as the measured bearing.
Indoors the signal arrives along many paths (Fig.~\ref{fig:wifi_habitat}), and when walls attenuate the direct path a reflection can be the strongest one, so the error of a single measurement is heavy-tailed~\cite{xiong2013arraytrack,kotaru2015spotfi,xie2019md}.
Separating the paths requires angular resolution, which grows with the array aperture~\cite{xiong2013arraytrack}; with the few antennas a commodity router exposes (four on ours), we therefore arrange them as a uniform linear array (ULA), which places the whole aperture on one axis.
Its cost is front--back ambiguity: the inter-antenna phase shift $\tfrac{2\pi d}{\lambda}\sin\theta$, with $\theta$ measured from the array's broadside, is identical for $\theta$ and $\pi-\theta$, so each measurement yields two world-frame bearing candidates $\beta_t^{\pm}$ (Fig.~\ref{fig:bearsolver}).
Both problems are addressed below by exploiting that \textit{a navigating robot, unlike a fixed receiver, sees the source from many vantage points along its trajectory.}

\textbf{Motion-based disambiguation.}
We exploit the fact that when the robot rotates in place by $\Delta\psi$ with the source fixed, the true candidate is invariant in the world frame while the mirror candidate rotates by $2\Delta\psi$.
In-place rotation is native to differential-drive platforms such as ours, and the module tolerates small drift: it buffers measurements taken during near-in-place rotations and, for each adjacent pair, matches the closest candidates across the pair; if they agree within $15^\circ$, a tolerance above the median AoA errors reported for commodity CSI systems~\cite{kotaru2015spotfi,xie2019md}, their mean is recorded as a \emph{verified} world-frame bearing.
No dedicated scanning motion is required: the turns the policy makes during navigation provide the rotations.

\textbf{Robust multi-view fusion.}
Verified bearings define rays $r_i=(\mathbf{p}_i,\mathbf{d}_i)$; before two verified bearings exist, both raw candidates of every measurement enter the ray set instead.
Because least-squares ray fitting is dominated by heavy-tailed outliers, we design a RANSAC-style~\cite{fischler1981random} multi-view estimator over bearing rays: each of $N_{\mathrm{iter}}{=}100$ iterations samples two rays, intersects them into a source hypothesis $\mathbf{g}$, and scores it by its number of inlier rays, i.e., rays passing within $\tau(\mathbf{g})$ of $\mathbf{g}$.
The threshold is range-adaptive, $\tau(\mathbf{g})=0.2\,\mathrm{m}+0.05\,\bar{d}_{\mathbf{g}}$ clamped to $[0.25,1]$\,m with $\bar{d}_{\mathbf{g}}$ the median distance from ray origins to $\mathbf{g}$, since a given bearing error translates into a larger position error for farther sources.
The current world-frame estimate $\hat{\mathbf{s}}_t$ is the best hypothesis with at least six inliers, refined as the mean of its projections onto the inlier rays.

\textbf{Convergence and confidence.}
The policy receives the running estimate at every step, tagged with a confidence $c_t=\max\big(1-e^{-n_v/10},\,0.2\,(1-e^{-n_r/100})\big)$ that grows quickly with the number $n_v$ of verified bearings and only slowly with the number $n_r$ of raw measurements; before the first estimate, the cue points $5$\,m ahead with $c_t{=}0$.
To prevent locking onto a premature estimate, it is frozen only once it is stable (mean displacement over the last $50$ updates below $0.1$\,m) and well supported (at least $50$ verified bearings), after which $c_t{=}1$.
The update costs $O(N_{\mathrm{iter}}|\mathcal{R}|)$ per step, negligible next to simulation and inference.

\begin{algorithm}[t]
\caption{Online Wi-Fi source localization}
\label{alg:loc}
\small
\DontPrintSemicolon
\KwIn{CSI $\mathbf{H}_t$, pose $(\mathbf{p}_t,\psi_t)$; raw history $\mathcal{H}_r$, verified history $\mathcal{H}_v$; previous estimate $\hat{\mathbf{s}}_{t-1}$}
\KwOut{cue $\mathbf{z}_t^w$}
\lIf{converged}{$\hat{\mathbf{s}}_t\leftarrow\hat{\mathbf{s}}_{t-1}$, $c_t\leftarrow1$, \textbf{go to} line~\ref{alg:loc:out}}
$\theta_t\leftarrow$ AoA of the dominant peak of the spectrum of $\mathbf{H}_t$\;
$\beta_t^{\pm}\leftarrow$ world-frame bearing candidates of $\theta_t$; append to $\mathcal{H}_r$\;
\tcp{motion-based disambiguation}
\ForEach{recent in-place rotation}{
  match candidates that stay fixed in the world frame\;
}
\lIf{a match exists}{append the verified bearing to $\mathcal{H}_v$}
\tcp{robust multi-view fusion}
$\mathcal{R}\leftarrow$ rays from $\mathcal{H}_v$ \textbf{if} $|\mathcal{H}_v|\ge2$ \textbf{else} both candidates in $\mathcal{H}_r$\;
\For{$N_{\mathrm{iter}}$ iterations}{
  intersect two sampled rays; keep the hypothesis with the largest inlier support\;
}
$\hat{\mathbf{s}}_t\leftarrow$ refine the best hypothesis on its inlier rays (else $\hat{\mathbf{s}}_{t-1}$)\;
update the converged flag and confidence $c_t$\;
\Return $\mathbf{z}_t^w$ from $\hat{\mathbf{s}}_t$ in the agent frame and $c_t$\label{alg:loc:out}\;
\end{algorithm}

\subsection{Localization-Aware Multimodal Policy and Training}
\label{sec:policy}
The policy (Fig.~\ref{fig:policy}) fuses the Wi-Fi cue with visual observations to select actions.
We use only the depth channel of the visual observation, encoded by a pretrained, frozen depth encoder: depth captures the geometry needed for obstacle avoidance and is less sensitive to sim-to-real appearance gaps than RGB.
The cue $\mathbf{z}_t^w$ is projected by a multi-layer perceptron (MLP), the previous action by a learned embedding, and the concatenated features drive a long short-term memory (LSTM) network followed by actor--critic heads.
Because the LSTM accumulates how the estimate has evolved, the policy is not bound to follow each intermediate estimate: it can keep moving, feeding the localizer new viewpoints, while the estimate is still being refined.

Training from scratch on ray-traced Wi-Fi is slow and unstable, since every step incurs ray tracing and early cues are noisy.
We therefore use a two-stage recipe: the policy is first pretrained with an analytic ground-truth AoA computed from the agent--source geometry in the same $\sin(\mathrm{AoA})$ form the AoA estimator produces, and then fine-tuned with CSI observations processed by the full localization module; only the input layer of the cue MLP is re-initialized in this switch, while the recurrent core and heads are carried over.
Both stages use a terminal success reward of $5$, a dense progress term equal to the per-step decrease in geodesic distance to the source, and a step penalty of $-0.01$.

%% file: Latex_Files/implementation.tex
\section{System Implementation}\label{sec:impl}

\begin{wrapfigure}{r}{0.46\linewidth}
    \vspace{-8pt}
    \centering
    \includegraphics[width=\linewidth]{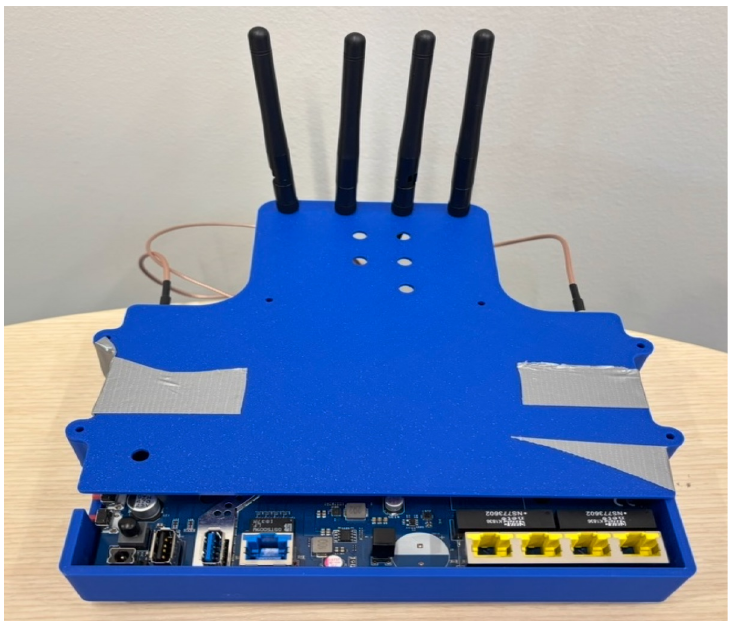}
    \caption{The router re-housed in a 3D-printed case as a four-antenna ULA.}
    \label{fig:wifi_hardware}
    \vspace{-8pt}
\end{wrapfigure}

\textbf{Real-robot platform.}
Our Wi-Fi receiver is an off-the-shelf ASUS RT-AC86U router~\cite{rtac86u} (Fig.~\ref{fig:wifi_hardware}).
We remove its enclosure and fix the four external antennas in a 3D-printed case as a ULA with $2.6\,\mathrm{cm}$ spacing, matching the simulated array geometry.
The router is flashed with Nexmon firmware~\cite{nexmon,gringoli2019free} to expose CSI from its commodity chipset, and CSI capture runs as ROS~2 nodes~\cite{hunter2023wiros}; CSI is phase-calibrated once with the WiROS procedure~\cite{hunter2023wiros}, which estimates a per-antenna, per-subcarrier phase-offset matrix against a transmitter at a known position.
The router runs in monitor mode as a CSI sniffer, overhearing frames from the target, selected by its MAC address, without connecting to it or its network; the target is a Motorola Edge 2020 smartphone that streams $1\,\mathrm{Mbit/s}$ of UDP traffic (iperf3, roughly 100 packets/s, a rate typical of a video call) to a laptop on the same Wi-Fi network.
Mounted on a Hello Robot Stretch~3, the array sits about $1.45\,\mathrm{m}$ above the ground (Fig.~\ref{fig:task}b); as in simulation, the localizer consumes one CSI measurement per policy step.
The policy runs on an NVIDIA Jetson AGX Orin; depth observations are denoised with Camera Depth Models~\cite{liu2025manipulation}, and discrete actions are executed under closed-loop control to limit drift. The pose $(\mathbf{p}_t,\psi_t)$ is the robot's onboard estimate (wheel odometry fused with 2D LiDAR scan matching) relative to the start pose, with no prior map or external localization.

\textbf{Simulation and training.}
\simname{} is configured to match this platform: a single-antenna transmitter, a $1\times4$ receiver ULA with $2.6\,\mathrm{cm}$ spacing, a $5.775\,\mathrm{GHz}$ carrier, $80\,\mathrm{MHz}$ bandwidth, and a maximum ray depth of 8, i.e., up to eight surface interactions per path, with transmitter and receiver placed $1.45\,\mathrm{m}$ above the floor; the agent body is $1.4\,\mathrm{m}$ tall with a $0.18\,\mathrm{m}$ radius.
Depth observations are rendered at $640\times480$ with a $58^\circ$ horizontal field of view (HFOV) and clipped to $[0.3, 3.0]\,\mathrm{m}$, matching the robot's depth camera, and the agent moves with a $0.25\,\mathrm{m}$ forward step and $30^\circ$ turns.
The depth encoder is a ResNet-50 pretrained on Habitat PointNav~\cite{wijmansdd} and kept frozen, and the recurrent core is an LSTM with hidden size 512.
Policies are trained with DD-PPO~\cite{wijmansdd} using VER~\cite{wijmans2022ver} on six NVIDIA L40S GPUs with 16 environment workers each, following the two-stage recipe of Sec.~\ref{sec:policy}: 60M steps of ground-truth-AoA pretraining take about 6 hours, and 12M steps of fine-tuning on ray-traced Wi-Fi observations take about 38 hours, the same ray-traced budget given to every policy in Table~\ref{tab:loc_ablation}. All policies are trained to convergence: training 25\% longer changes success rate by about one point.

%% file: Latex_Files/experiments.tex
\section{Experiments}\label{sec:exp}
\subsection{Experimental Setup}
\label{sec:exp_setup}
We train and evaluate on visual--Wi-Fi navigation episodes (Sec.~\ref{sec:task}) built with Habitat's standard episode sampler~\cite{savva2019habitat}: each episode draws a navigable start pose and a target location on the navigation mesh, keeping pairs whose geodesic-to-Euclidean distance ratio is at least $1.1$ (no straight-line episodes; test-set median $1.24$), and the target location becomes the Wi-Fi transmitter position; the source is out of sight from $95.1\%$ of start poses. We use the official scene splits of HSSD~\cite{khanna2024habitat} (122 training and 42 test scenes) and sample 30 episodes per test scene (1{,}260 test episodes), and additionally evaluate the HSSD-trained policy zero-shot on the test scenes of ReplicaCAD~\cite{szot2021habitat} and ProcTHOR~\cite{ProcTHOR} (Sec.~\ref{sec:cross_dataset}).
All simulated experiments use \simname{} with the setup of Sec.~\ref{sec:impl}; in simulation the agent pose is provided by Habitat.
Following prior embodied navigation works~\cite{wijmans2022ver,shi2025towards,long2025instructnav}, we report Success Rate (SR), Success weighted by Path Length (SPL), and Distance-to-Goal (DTG) with a success distance of $0.5\,\mathrm{m}$, and additionally the localization error of the online source estimate at the final step, which reflects both the localizer and the agent's trajectory.
\subsection{\sysname{} Learns Effective Navigation in \simname{}}
\begin{table}[t]
\centering
\footnotesize
\caption{
Main results on HSSD test scenes.
GT: ground truth; oracle: GT goal cue; Wi-Fi Obs.: CSI simulated by \simname{}; Est. Goal: estimate of our online localization module; w/o FT: no fine-tuning on Wi-Fi observations. SR, SPL in percent; DTG, Loc. in meters.
}
\label{tab:main_results}
\setlength{\tabcolsep}{4pt}
\begin{tabular}{llcccc}
\toprule
Method & Goal Cue & SR $\uparrow$ & SPL $\uparrow$ & DTG $\downarrow$ & Loc. $\downarrow$ \\
\midrule
No-goal \sysname{} & -- & 0.16 & 0.10 & 10.27 & -- \\
PointNav (oracle) & GT PointGoal & 87.22 & 69.88 & 1.66 & -- \\
\sysname{} (oracle) & GT AoA & 82.30 & 60.83 & 2.10 & $<\!10^{-3}$ \\
\midrule
PointNav & Est. Goal & 49.76 & 34.57 & 4.99 & 3.23 \\
Loc-then-PointNav & Est. Goal & 17.78 & 8.91 & 6.50 & 2.85 \\
\sysname{} (w/o FT) & Wi-Fi Obs. & 53.02 & 35.82 & 4.58 & 2.56 \\
\sysname{} & Wi-Fi Obs. & \textbf{68.57} & \textbf{44.12} & \textbf{3.04} & \textbf{1.24} \\
\bottomrule
\end{tabular}
\end{table}

We compare three navigation strategies and a no-goal control (Table~\ref{tab:main_results}); PointNav and \sysname{} each appear in an oracle setting that supplies a ground-truth (GT) goal cue and in the realistic setting where the cue must be recovered from Wi-Fi.
\textbf{PointNav} is a standard point-goal policy that takes a target vector as input, trained by us on the same HSSD episodes and embodiment with GT goals and used off the shelf. Given the GT target vector it is an oracle upper bound. Replacing the GT target vector with the estimate of our online localization module, the same policy tests whether the task can be solved by reusing PointNav directly.
\textbf{Loc-then-PointNav} first localizes the source by random exploration until localization convergence or 100 steps, then runs PointNav to the frozen estimate, i.e., the frozen-estimate counterpart of \sysname{}.
Both are \emph{decoupled}: a point-goal policy trained on ground-truth goals consumes an estimate it was never trained on.
\textbf{\sysname{}} is our closed-loop framework, pretrained on GT AoA and fine-tuned on Wi-Fi observations in \simname{}; (oracle) is its upper bound with GT AoA, and (w/o FT) skips the fine-tuning.
\textbf{No-goal \sysname{}} is a control receiving no wireless cue, only depth and the previous action.

\begin{figure}[t]
    \centering
    \includegraphics[width=\linewidth]{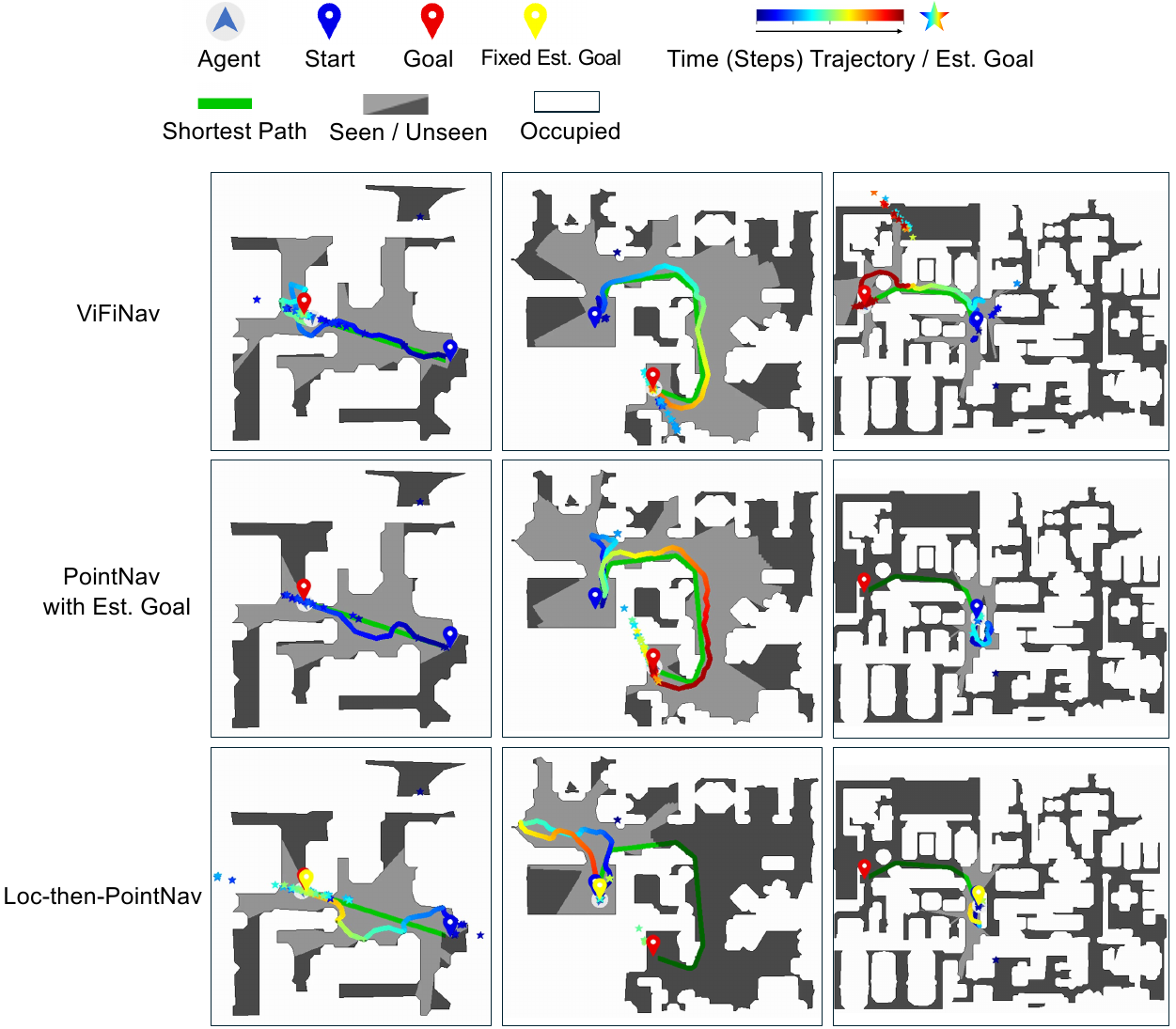}
    \caption{
    Qualitative comparison on three HSSD test episodes (columns). Rows: \sysname{} (top), PointNav (middle), Loc-then-PointNav (bottom). Stars mark successive source estimates, coloured by time. \sysname{}'s estimates migrate toward the source as it moves, keeping its path short. PointNav follows each updated estimate and detours. Loc-then-PointNav freezes one early estimate and cannot recover when it is wrong.
    }
    \label{fig:qual_example_grid}
\end{figure}

No-goal \sysname{} fails ($0.16\%$ SR), confirming that the episodes cannot be solved by exploration or obstacle avoidance alone.
With oracle cues, PointNav (oracle) reaches $87.22\%$ SR and \sysname{} (oracle) $82.30\%$, showing that accurate directional wireless cues suffice for this task.
\sysname{} (oracle) trails PointNav (oracle) because its GT AoA is front--back-ambiguous and carries no range (Sec.~\ref{sec:policy}), so the localization module must still resolve it through motion; its sub-millimetre localization error is the numerical residual of fusing exact bearings.
Without fine-tuning, however, \sysname{} (w/o FT) drops to $53.02\%$ SR on ray-traced Wi-Fi observations, and the modular baselines fare no better (PointNav $49.76\%$, Loc-then-PointNav $17.78\%$).
Fine-tuning on Wi-Fi observations recovers much of this gap: \sysname{} reaches $68.57\%$ SR and $44.12\%$ SPL, and its trajectories also yield better online source estimates ($2.56$\,m to $1.24$\,m final localization error).
Trained only on GT cues and tested on the estimate, \sysname{} (w/o FT) already edges out PointNav ($53.02\%$ vs.\ $49.76\%$); the larger gain comes from closed-loop fine-tuning on Wi-Fi observations, which \simname{} makes possible and which is part of \sysname{} itself (Fig.~\ref{fig:policy}).
Rather than treating localization and navigation as separate stages, \sysname{} learns to act on evolving source estimates, so that its motion improves the quality of future localization cues; Fig.~\ref{fig:qual_example_grid} illustrates this on three test episodes, where PointNav chases every updated estimate and Loc-then-PointNav commits to an early inaccurate one, while \sysname{} keeps exploring until its estimate converges near the source.

\subsection{Component Analysis of \sysname{}}

\begin{table}[t]
\centering
\footnotesize
\caption{Effect of the fusion estimator, ambiguity resolution, and two-stage training. LS: least squares.}
\label{tab:loc_ablation}
\setlength{\tabcolsep}{3pt}
\resizebox{\columnwidth}{!}{%
\begin{tabular}{lcccccc}
\toprule
Loc. Alg. & Amb. Resolver & Pretrain & SR $\uparrow$ & SPL $\uparrow$ & DTG $\downarrow$ & Loc. Err. $\downarrow$ \\
\midrule
LS & -- & --  & 1.75 & 1.30 & 7.84 & 48.74 \\
LS & \checkmark & --  & 43.33 & 28.14 & 4.91 & 4.34 \\
RANSAC & -- & -- & 47.62 & 27.95  & 4.70 & 2.03 \\
RANSAC & \checkmark & -- & 52.22 & 30.17  & 4.20 & 1.26 \\
RANSAC & \checkmark & \checkmark & 68.57 & 44.12 & 3.04 & 1.24 \\
\bottomrule
\end{tabular}%
}
\end{table}

Table~\ref{tab:loc_ablation} varies three design choices: how bearing rays are fused, whether front--back ambiguity is resolved before fusion, and whether the policy is trained directly on ray-traced Wi-Fi or with the two-stage recipe of Sec.~\ref{sec:policy}. LS fits the source position to all bearing rays by least squares with no outlier rejection; RANSAC is the estimator of Sec.~\ref{sec:loc}, which keeps the hypothesis with the largest inlier support.
\textbf{Ambiguity resolution} is indispensable for LS, lifting SR from $1.75\%$ to $43.33\%$ and cutting localization error from $48.74\,\mathrm{m}$ to $4.34\,\mathrm{m}$: unresolved candidates otherwise place half the rays in the mirrored direction.
\textbf{Robust fusion} reaches much of the same effect on its own, since RANSAC can treat those mirrored rays as outliers ($47.62\%$ SR, $2.03\,\mathrm{m}$ without the resolver), and the two together are best ($52.22\%$, $1.26\,\mathrm{m}$).
\textbf{Two-stage training} adds the largest remaining gain, $52.22\%\!\rightarrow\!68.57\%$ SR and $30.17\%\!\rightarrow\!44.12\%$ SPL, while final localization error barely moves ($1.26$ to $1.24\,\mathrm{m}$): the gain lies in policy training, not in the estimator.

\begin{figure}[t]
    \centering
    \begin{subfigure}[b]{0.48\linewidth}
        \centering
        \includegraphics[width=\linewidth]{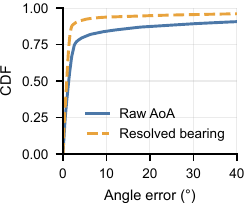}
        \caption{Direction error.}
        \label{fig:loc_analysis_a}
    \end{subfigure}
    \hfill
    \begin{subfigure}[b]{0.48\linewidth}
        \centering
        \includegraphics[width=\linewidth]{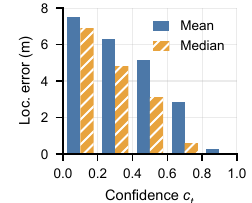}
        \caption{Error vs.\ confidence.}
        \label{fig:loc_analysis_b}
    \end{subfigure}
    \caption{
    Localization module analysis on $10{,}000$ HSSD training-split episodes, replayed along their shortest paths. (a) Cumulative distribution function (CDF) of direction error before (raw AoA) and after (resolved bearing) motion-based ambiguity resolution. (b) Mean and median source localization error per confidence bin.
    }
    \label{fig:loc_analysis}
\end{figure}
Fig.~\ref{fig:loc_analysis} examines the localization module on its own. Since the module is analytic and has no learned parameters, we evaluate it on $10{,}000$ HSSD training-split episodes, replayed along their shortest paths so that the viewpoints are fixed rather than policy-dependent.
Raw AoA estimates are accurate for most observations (median $1.3^\circ$) but heavy-tailed (mean $12.2^\circ$); the motion-based ambiguity resolver reduces the mean error to $5.1^\circ$ and removes most of the tail (Fig.~\ref{fig:loc_analysis_a}), which explains why unresolved AoA fails in Table~\ref{tab:loc_ablation}.
The confidence $c_t$ is also informative: mean localization error drops from $7.5\,\mathrm{m}$ in the lowest confidence bin to $0.3\,\mathrm{m}$ in the highest (Fig.~\ref{fig:loc_analysis_b}).

\subsection{Cross-Dataset Generalization}
\label{sec:cross_dataset}

\begin{table}[t]
\centering
\footnotesize
\caption{Zero-shot cross-dataset evaluation of the HSSD-trained \sysname{}, with test-set statistics (scenes, episodes, mean geodesic start--goal distance).}
\label{tab:cross_dataset}
\setlength{\tabcolsep}{3.5pt}
\resizebox{\columnwidth}{!}{%
\begin{tabular}{lccc cccc}
\toprule
Dataset & Scenes & Episodes & Geo. dist.\ (m) & SR$\uparrow$ & SPL$\uparrow$ & DTG$\downarrow$ & Loc.$\downarrow$ \\
\midrule
HSSD (in-domain) & 42 & 1{,}260 & 11.30 & 68.57 & 44.12 & 3.04 & 1.24 \\
ReplicaCAD & 21 & 1{,}050 & 7.44 & 75.05 & 47.14 & 1.29 & 0.72 \\
ProcTHOR & 1{,}000 & 5{,}000 & 7.19 & 85.22 & 50.83 & 1.29 & 0.20 \\
\bottomrule
\end{tabular}%
}
\end{table}

As shown in Table~\ref{tab:cross_dataset}, we test the HSSD-trained \sysname{} on episodes sampled as in Sec.~\ref{sec:exp_setup} on the official held-out ReplicaCAD and ProcTHOR scenes, without fine-tuning.
\sysname{} achieves $75.05\%$ SR on ReplicaCAD and $85.22\%$ SR on ProcTHOR, suggesting that the learned visual--Wi-Fi policy transfers to scene families beyond HSSD.
This transfer is consistent with our modality design: depth observations reduce sensitivity to dataset-specific appearance, while Wi-Fi cues depend primarily on scene geometry and the source--agent configuration.
The higher absolute SR on the unseen datasets partly reflects their shorter geodesic distances.

\subsection{Comparison with Object-Goal Navigation}
\label{sec:objectnav}

\begin{table}[t]
\centering
\footnotesize
\caption{
ObjectNav policies vs.\ \sysname{} on HSSD ObjectNav episodes.
Category-level metrics count any instance of the goal category, instance-level metrics only the designated instance, at which the Wi-Fi source is placed for \sysname{}; success distance $0.5\,\mathrm{m}$ to the instance viewpoints at both levels.
``Official'' follows the HSSD recipe~\cite{khanna2024habitat}; ``matched'' uses the embodiment of \sysname{} (Sec.~\ref{sec:impl}).
}
\label{tab:objectnav}
\setlength{\tabcolsep}{4pt}
\resizebox{\columnwidth}{!}{%
\begin{tabular}{lccc ccc}
\toprule
& \multicolumn{3}{c}{Category-level} & \multicolumn{3}{c}{Instance-level} \\
\cmidrule(lr){2-4} \cmidrule(lr){5-7}
Method & SR $\uparrow$ & SPL $\uparrow$ & DTG $\downarrow$ & SR $\uparrow$ & SPL $\uparrow$ & DTG $\downarrow$ \\
\midrule
ObjectNav (official) & 60.95 & 28.38 & 2.68 & 25.67 & 13.20 & 7.55 \\
ObjectNav (matched) & 61.45 & 27.26 & 2.84 & 25.07 & 12.86 & 7.69 \\
\sysname{} (zero-shot) & -- & -- & -- & \textbf{56.29} & \textbf{30.40} & \textbf{4.59} \\
\bottomrule
\end{tabular}%
}
\end{table}

In practice, a user often wants a \emph{specific} object, e.g., their own phone rather than any phone.
Object-goal navigation (ObjectNav) specifies the target by category and therefore cannot distinguish instances, whereas a Wi-Fi target is identified by its MAC address, so one instance is distinguished from others of the same category.
To quantify this, we evaluate on the official HSSD ObjectNav validation episodes~\cite{khanna2024habitat}, keeping the 1{,}009 episodes whose designated instance is reachable under both embodiments, and score each episode both at the category level (stopping near any instance of the goal category) and at the instance level, where one instance is designated as the target.
For \sysname{}, the Wi-Fi source is placed at a navigable viewpoint of the designated instance, and we run the checkpoint of Table~\ref{tab:main_results} zero-shot.
We compare against two ObjectNav policies that we train ourselves on the HSSD training split.\footnote{The released HSSD checkpoint (huggingface.co/datasets/hssd/hssd-checkpoints) was selected for transfer to HM3D.} The \emph{official} policy follows the HSSD recipe~\cite{khanna2024habitat}; the \emph{matched} policy differs only in embodiment, which we match to \sysname{} (four actions, $1.4\,\mathrm{m}$ agent height, $58^\circ$ HFOV, RGB input), so that any gap can be attributed to the goal specification alone.

At the category level, the official policy reaches $60.95\%$ SR ($55.00\%$ under the standard $0.1\,\mathrm{m}$ criterion, matching the $54.8\%$ reported in~\cite{khanna2024habitat}).
Once a specific instance is designated, however, SR drops by more than half, to $25.67\%$, and the mean distance to the goal grows from $2.68\,\mathrm{m}$ to $7.55\,\mathrm{m}$: the policy reliably stops next to \emph{an} object of the right category, but has no cue to prefer the designated one.
The matched policy behaves the same ($25.07\%$).
\sysname{}, given no semantic description of the target, reaches the designated instance in $56.29\%$ of episodes with $30.40\%$ SPL, more than doubling both, because the Wi-Fi source resolves the target by localization rather than recognition.
Its SR is below Table~\ref{tab:main_results} because these episodes are out of distribution.

\subsection{Real-World Deployment}

\begin{figure}[t]
    \centering
    \begin{subfigure}[b]{0.32\linewidth}
        \centering
        \includegraphics[width=\linewidth]{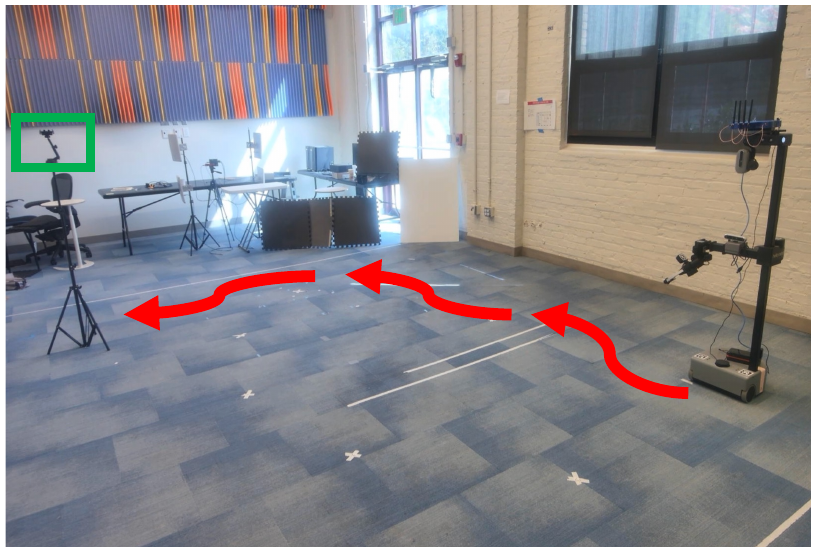}
        \caption{LoS.}
    \end{subfigure}
    \hfill
    \begin{subfigure}[b]{0.32\linewidth}
        \centering
        \includegraphics[width=\linewidth]{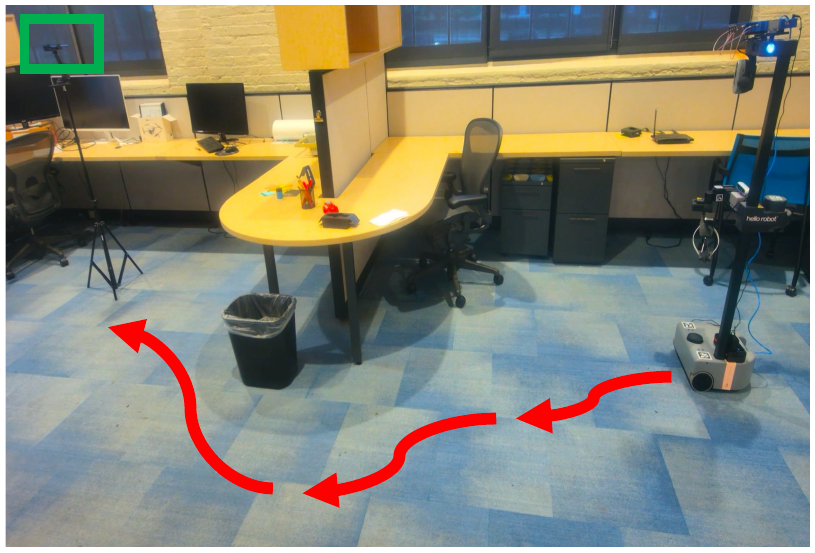}
        \caption{NLoS.}
    \end{subfigure}
    \hfill
    \begin{subfigure}[b]{0.32\linewidth}
        \centering
        \includegraphics[width=\linewidth]{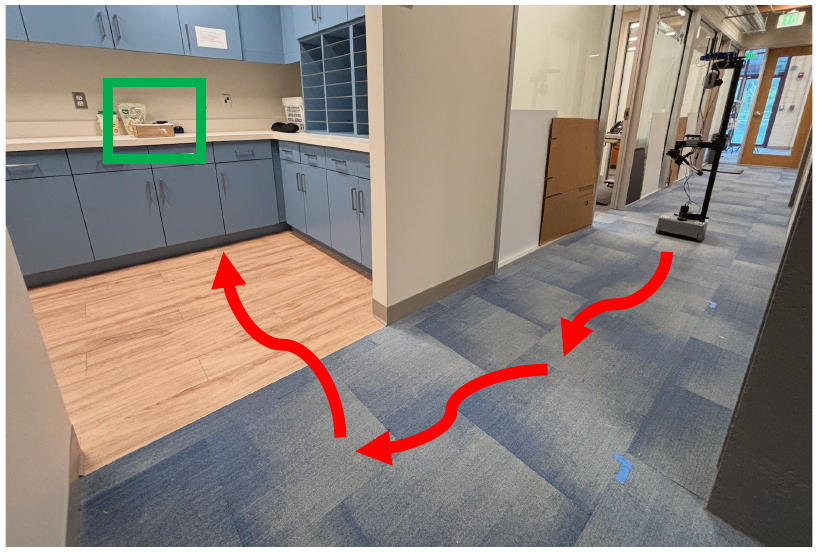}
        \caption{Applications.}
    \end{subfigure}
\caption{
Representative real-world trials: (a) line-of-sight, (b) non-line-of-sight, and (c) an application scenario in which the Wi-Fi source (a phone) is inside a package. Green box: Wi-Fi source; red: robot trajectory.
}
\label{fig:realsetup}
\end{figure}

Finally, we deploy \sysname{} on the robot platform of Sec.~\ref{sec:impl}, using the simulation-trained policy without any real-world fine-tuning.
A trial succeeds when the robot stops within $1\,\mathrm{m}$ of the Wi-Fi device, and terminates upon success, collision, or 200 steps; the radius is relaxed from the $0.5\,\mathrm{m}$ used in simulation to account for odometry drift and the target's physical extent, and matches the $1\,\mathrm{m}$ standard in real-world navigation~\cite{gervet2023navigating}.
Trials are conducted in an office space with drywall and glass partitions, metal cabinets, and wooden furniture; in NLoS trials the source is placed behind fabric-covered cubicle partitions or an interior drywall wall.
We evaluate in line-of-sight (LoS) and non-line-of-sight (NLoS) settings (Fig.~\ref{fig:realsetup}a,b), with 10 trials each over varied start poses and source placements; \sysname{} succeeds in 7 out of 10 LoS and 5 out of 10 NLoS trials. 
Over the 20 trials, the final error of the online source estimate averages $1.16$\,m in LoS and $1.92$\,m in NLoS ($1.54$\,m overall), against $1.24$\,m for the simulated episodes of Table~\ref{tab:main_results}.
We further demonstrate application scenarios from Fig.~\ref{fig:task}, e.g., package search with a phone inside a package (Fig.~\ref{fig:realsetup}c). Example trial videos are provided in the supplementary material.

%% file: Latex_Files/conclusion.tex
\section{Conclusion and Future Work}
\label{sec:conclusion}
We presented \simname{}, an embodied navigation simulator with online ray-traced Wi-Fi sensing, and \sysname{}, a localization-aware visual--Wi-Fi navigation framework trained in it. In simulation, fine-tuning on ray-traced Wi-Fi observations improves success rate, and the policy transfers zero-shot to unseen scene datasets. On a real robot with a Wi-Fi sensor built from a commodity router, the same policy is deployed without any real-world training.
The approach is not tied to this hardware: only the localization front-end is Wi-Fi specific, and \simname{} exposes carrier, bandwidth, array geometry, and materials as configuration, so UWB or BLE direction finding could replace Wi-Fi.
Future work includes randomizing surface materials in training and extending online ray tracing to scanned scenes to narrow the sim-to-real gap, and evaluating the estimator under sparse or bursty target traffic. Learning the Wi-Fi processing jointly with the policy could replace the analytic front-end, and vision--language models over visual and Wi-Fi cues could admit language-specified goals.